\documentclass{article}

 \usepackage[preprint]{neurips_2025}

 \workshoptitle{Published in the proceedings of the 39th Conference on Neural Information Processing Systems (NeurIPS 2025) Workshop: Symmetry and Geometry in Neural Representations (NeurReps). Additionally accepted for presentation in NeurIPS 2025 Workshop: Interpreting Cognition in Deep Learning Models (CogInterp).}

\usepackage[utf8]{inputenc} 
\usepackage[T1]{fontenc}    
\usepackage[colorlinks=true, allcolors=blue]{hyperref}
\usepackage{url}            
\usepackage{booktabs}       
\usepackage{amsfonts}       
\usepackage{nicefrac}       
\usepackage{microtype}      
\usepackage{xcolor}         
\usepackage{amsmath}
\usepackage[pdftex]{graphicx}
\usepackage{wrapfig}

\title{The Geometry of Cortical Computation: Manifold Disentanglement and Predictive Dynamics in VCNet}

\title{The Geometry of Cortical Computation: Manifold Disentanglement and Predictive Dynamics in VCNet}

\author{%
Brennen Hill\thanks{Corresponding author.} \\
Department of Computer Science\\
University of Wisconsin-Madison\\
Madison, WI 53706 \\
\texttt{bahill4@wisc.edu} \\
\And
Zhang Xinyu \\
Department of Computer Science\\
National University of Singapore\\
Singapore 119077 \\
\texttt{zhang\_xinyu\_@u.nus.edu} \\
\And
Timothy Putra Prasetio \\
Department of Computer Science\\
National University of Singapore\\
Singapore 119077 \\
\texttt{timothy\_prasetio@u.nus.edu} \\
}

\begin{document}

\maketitle

\begin{abstract}
Despite their success, modern convolutional neural networks (CNNs) exhibit fundamental limitations, including data inefficiency, poor out-of-distribution generalization, and vulnerability to adversarial perturbations. These shortcomings can be traced to a lack of inductive biases that reflect the inherent geometric structure of the visual world. The primate visual system, in contrast, demonstrates superior efficiency and robustness, suggesting that its architectural and computational principles,which evolved to internalize these structures,may offer a blueprint for more capable artificial vision. This paper introduces Visual Cortex Network (VCNet), a novel neural network architecture whose design is informed by the macro-scale organization of the primate visual cortex. VCNet is framed as a geometric framework that emulates key biological mechanisms, including hierarchical processing across distinct cortical areas, dual-stream information segregation for learning disentangled representations, and top-down predictive feedback for representation refinement. We interpret these mechanisms through the lens of geometry and dynamical systems, positing that they guide the learning of structured, low-dimensional neural manifolds. We evaluate VCNet on two specialized benchmarks: the Spots-10 animal pattern dataset, which probes sensitivity to natural textures, and a light field image classification task, which requires processing higher-dimensional visual data. Our results show that VCNet achieves state-of-the-art accuracy of 92.1\% on Spots-10 and 74.4\% on the light field dataset, surpassing contemporary models of comparable size. This work demonstrates that integrating high-level neuroscientific principles, viewed through a geometric lens, can lead to more efficient and robust models, providing a promising direction for addressing long-standing challenges in machine learning.
\end{abstract}

\section{Introduction}

Contemporary deep learning models for image recognition, particularly Convolutional Neural Networks (CNNs), have achieved remarkable success \cite{krizhevsky2012imagenet}. However, their prowess is shadowed by critical challenges that impede their widespread and reliable deployment. These models often require vast, meticulously labeled training datasets, exhibit poor generalization to out-of-distribution (OOD) examples \cite{sagawa2020distributionally}, and are notoriously vulnerable to adversarial attacks and partial occlusion \cite{liu2022deep}. Minor, human-imperceptible perturbations or hidden object parts can cause catastrophic failures, raising concerns about their reliability in safety-critical applications. These persistent issues are not merely implementation details but may point to a fundamental inadequacy in their core architectural assumptions. While CNNs incorporate a crucial geometric prior-translation equivariance,they largely fail to account for other fundamental symmetries and structures of the visual world, such as rotation, scale, and the compositional nature of objects.

In stark contrast, the primate visual system is a paragon of efficiency and robustness. Humans can learn to recognize objects from few examples \cite{Lake2015:HumanLevel}, generalize effortlessly across novel contexts \cite{Geirhos2018:Generalisation}, robustly identify occluded objects \cite{Hegde2008:Preferential}, and operate with unparalleled energy efficiency \cite{lennie2003cost}. Mounting evidence from neuroscience suggests these capabilities are rooted in the specific architectural and computational principles of the visual cortex. Neural representations in the brain appear to be organized on low-dimensional, structured geometric spaces, often referred to as neural manifolds. The brain seems to learn not just features, but the underlying geometric and topological structure of the data-generating process. This is achieved through its unique hierarchical organization \cite{Felleman1991:Distributed, GrillSpector2004:HumanVisual} and its use of predictive processing \cite{Rao1999:Predictive, deLange2018:Expectations}.

This paper seeks to bridge the gap between the brute-force pattern matching of modern CNNs and the geometrically-aware, structured inference of the brain. We propose VCNet, a novel neural network whose macro-architecture is derived from the primate visual cortex. We go beyond mere biomimicry and interpret the cortex's organization as a computational framework for learning geometrically sound representations. Our work is guided by the thesis that principles of symmetry and geometry can illuminate the foundations of intelligence. Our contributions are threefold:
\begin{itemize}
    \item We introduce \textbf{VCNet}, a deep neural network architecture that models the high-level information flow between major areas of the visual cortex. We provide a \textbf{geometric interpretation} of its core components, including dual-stream processing for manifold disentanglement, recurrent connections for representation dynamics, and top-down predictive feedback as a mechanism for geometric refinement on these manifolds.
    \item We demonstrate the efficacy of VCNet on the \textbf{Spots-10 animal pattern benchmark}. This task is selected to test our bio-inspired architecture on a problem that mirrors a key evolutionary pressure for vision, and we show that VCNet outperforms other models of comparable size.
    \item We further evaluate VCNet on a \textbf{light field image classification task}, providing evidence that its geometrically-motivated design is particularly well-suited for processing richer, multi-view data that more closely approximates the input to the human visual system.
\end{itemize}

\section{Related Work}

Our research is situated at the confluence of geometric deep learning, computational neuroscience, and neuro-inspired AI.

\paragraph{Geometric Deep Learning and Equivariance}
Geometric deep learning seeks to incorporate geometric priors and symmetries into neural network architectures \cite{Bronstein2021:GeometricDeep}. A significant focus has been on achieving equivariance to groups of transformations beyond translation. Steerable CNNs \cite{Cohen2016:Steerable} and E(2)-equivariant CNNs \cite{Weiler2019:GeneralE2} generalize convolutions to handle rotation and reflection, leading to improved data efficiency and generalization on tasks with these underlying symmetries. Graph Neural Networks (GNNs) extend these ideas to data defined on non-Euclidean domains like graphs and manifolds. While these approaches enforce geometric constraints at the micro-level of individual filters or operations, VCNet takes a complementary, macro-level approach. We do not engineer specific filter symmetries; instead, we hypothesize that the high-level architectural organization of the visual cortex itself, its network topology of processing streams and feedback loops, creates an inductive bias that guides the learning process toward geometrically structured representations.

\paragraph{Neuro-Inspired Architectures}
The brain has long been a source of inspiration for artificial intelligence. Early models like the Neocognitron \cite{Fukushima1980:Neocognitron} laid the groundwork for modern CNNs by mimicking the simple and complex cells of V1. More recently, models like CorNet \cite{Kubilius2019:BrainLike} have sought to create architectures that not only perform well but also whose internal activations correlate with neural recordings from the primate brain. These models often focus on replicating the feedforward ventral stream. VCNet differentiates itself by modeling a more comprehensive set of cortical principles: (1) the explicit separation and interaction of the ventral and dorsal streams, (2) the inclusion of recurrent dynamics to model iterative processing, and (3) the implementation of a top-down predictive coding loop, which we argue is critical for robust, generative understanding.

\paragraph{Predictive Coding and Generative Models}
Predictive coding posits that the brain is fundamentally a generative model of its environment \cite{Rao1999:Predictive}. Higher cortical areas generate predictions about lower-level sensory input, and only the residual error between the prediction and the actual input is propagated forward. This principle is computationally efficient and has deep connections to Bayesian inference and the free-energy principle \cite{Friston2010:FreeEnergy}. In machine learning, this resonates with the objectives of generative models like Variational Autoencoders (VAEs) and Helmholtz machines, which learn a latent generative model of the data. Our implementation of predictive coding in VCNet serves a similar purpose, encouraging the network to learn an internal model of the visual world. By framing this process geometrically, we view the prediction error as a vector in the tangent space of a learned representation manifold, driving the model's state along a geodesic toward a more accurate representation.

\section{The VCNet Architecture: A Geometric Interpretation}

While a complete replication of the visual system is infeasible, our research focuses on emulating the macro-scale organization of the visual cortex. We interpret its connectivity and computational patterns not as an arbitrary biological arrangement, but as a sophisticated framework for learning and processing geometric information. VCNet is a deep neural architecture engineered to operationalize these principles.

\subsection{Biologically-Inspired Geometric Principles}

Our model's design is predicated on three foundational principles of primate vision, which we reformulate in the language of geometry and dynamics.

\paragraph{Hierarchical Processing as Compositional Feature Geometry}
Visual information propagates from the retina through a hierarchy of cortical areas (V1, V2, V3, V4, V5), each specialized for extracting progressively complex features \cite{Huff2023:Neuroanatomy}. V1 detects simple elements like oriented edges, which can be understood as recognizing local Euclidean symmetries (translations and rotations of local patterns). It projects to V2, which processes intermediate features like contours and textures. V2, in turn, projects to higher-order areas like V4 and V5. We interpret this hierarchy not merely as a cascade of feature extractors, but as a sequence of learned, non-linear transformations $\phi_i: \mathcal{M}_{i-1} \to \mathcal{M}_i$ that map representations from one manifold $\mathcal{M}_{i-1}$ to another $\mathcal{M}_i$. Each stage aims to create a new representation that is more abstract, more disentangled, and more useful for the organism's goals. The final representation should ideally live on a manifold where semantic categories are linearly separable.

\begin{figure}[h]
\centering
\includegraphics[width=0.8\linewidth]{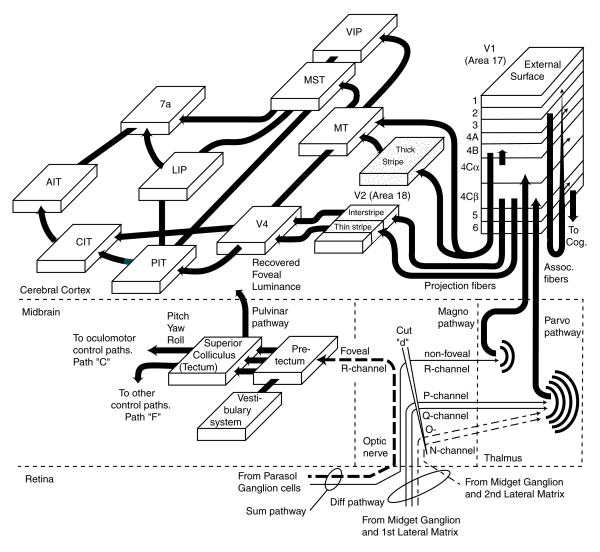}
\caption{A high-level model of information pathways in the primate visual cortex, illustrating the hierarchical series of feature extraction stages \cite{fulton2001processes}. This organization forms the architectural basis of VCNet, which we interpret as a graph of transformations between representation manifolds.}
\label{fig:cortex_diagram}
\end{figure}

\paragraph{Dual-Stream Processing as Manifold Disentanglement}
The visual cortex is famously organized into two primary processing pathways \cite{Sheth2016:TwoVisual}. The \textbf{ventral stream} ("what" pathway) is responsible for object recognition, while the \textbf{dorsal stream} ("where/how" pathway) handles spatial awareness and motion analysis. We propose a geometric interpretation: these two streams learn to project the high-dimensional sensory input onto two distinct, yet correlated, low-dimensional manifolds.
\begin{itemize}
    \item The \textbf{ventral stream} learns an \textbf{identity manifold}, $\mathcal{M}_{id}$. The goal is to learn a representation that is \textit{invariant} to changes in pose, illumination, and position. Points on this manifold correspond to object identities.
    \item The \textbf{dorsal stream} learns a \textbf{pose/motion manifold}, $\mathcal{M}_{pose}$. The goal here is to learn a representation that is \textit{equivariant} to changes in object position and orientation. Points on this manifold correspond to spatial properties.
\end{itemize}
By explicitly separating these tasks into different architectural pathways, the model is encouraged to learn disentangled representations, a key goal in representation learning. The interconnections between the streams allow the model to bind "what" information with "where" information.

\paragraph{Predictive Coding as Geodesic Refinement on a Manifold}
The visual cortex is not a purely feedforward system. It employs predictive coding, where higher-level areas send top-down predictions to lower-level areas \cite{Lowet2024:Distinction}. We formalize this as a process of refinement on a representation manifold. Let $z_L$ be a representation in a high-level area (e.g., AIT). The network learns a top-down mapping $p(z_l | z_L)$ that generates a prediction of the representation $z_l$ in a lower-level area (e.g., V1). The bottom-up processing provides the actual sensory evidence, resulting in an observed representation $z_l^{obs}$. The discrepancy, or prediction error $\epsilon$, is computed:
$$ \epsilon = z_l^{obs} - p(z_l | z_L) $$
This error signal is not just noise; it is a vector in the tangent space of the lower-level manifold $\mathcal{M}_l$. This vector indicates the direction in which the higher-level representation $z_L$ needs to be updated to better explain the sensory data. The learning process, which aims to minimize this error over time and data, can be viewed as an optimization process that encourages the model's internal generative trajectory to follow a geodesic on the manifold of plausible world states. This creates a powerful dynamical system for inference and learning.

\subsection{Architectural Framework of VCNet}

Departing from conventional, monolithic CNN architectures, VCNet is structured as a directed graph that models the known connectivity between the major visual cortical areas. The channel capacity of each module is scaled to approximate the relative neuronal populations in its biological counterpart.

\paragraph{Ventral Stream} This "what" pathway models object recognition, progressing from V1 through modules representing V2 (Interstripe, Thin Stripe), V4, and the inferotemporal (PIT, CIT, AIT) cortices. It is specialized for learning the invariant \textit{identity manifold}.

\paragraph{Dorsal Stream} This "where/how" pathway models spatial and motion analysis, flowing from V1 through V2 (Thick Stripe), the middle temporal (MT) and medial superior temporal (MST) areas, and onward to parietal regions. It is specialized for learning the equivariant \textit{pose/motion manifold}.

These streams are interconnected at multiple levels, enabling the integration of object identity with spatial information. The final representation is formed in the AIT module, which receives convergent inputs and feeds into the classification layer. VCNet's functionality is realized through several specialized computational blocks, each with a geometric interpretation.

\subsubsection{Multi-Scale Feature Extraction (V1)}
\textbf{Implementation:} To emulate the diverse receptive field sizes in V1, the module processes input through three parallel depthwise separable convolution streams with different kernel sizes (3x3, 5x5, 7x7). The resulting feature maps are concatenated.
\textbf{Geometric Interpretation:} This block acts as a multi-scale probe of the local geometry of the input signal. Different kernel sizes are sensitive to structures at different frequencies and scales, akin to a wavelet decomposition. This provides a rich, multi-scale initial representation that captures the geometry of the input space more effectively than a single-scale approach.

\subsubsection{Recurrent Processing Blocks (MT/MST)}
\textbf{Implementation:} To model the iterative refinement of representations, the MT and MST modules incorporate Recurrent Blocks. These blocks apply a convolutional transformation with shared weights for a fixed number of iterations ($t=3$), with each iteration receiving the output of the previous one plus a residual connection from the initial input.
\textbf{Geometric Interpretation:} This recurrent application of a transformation defines a discrete-time dynamical system on the feature space. The representation $z_t$ at iteration $t$ evolves according to $z_{t+1} = f(z_t) + z_0$. The recurrence allows the representation to iteratively converge toward a stable fixed point on its manifold, effectively refining the estimate of motion or spatial properties over time.

\subsubsection{Attentional Modulation (CBAM)}
\textbf{Implementation:} To emulate the brain's ability to focus on salient features, key modules (V1, MT, V4) incorporate a Convolutional Block Attention Module (CBAM). CBAM sequentially infers and applies channel-wise and spatial attention maps.
\textbf{Geometric Interpretation:} Attention can be viewed as a mechanism for dynamically selecting a relevant subspace of the feature manifold. Channel attention re-weights the contribution of different feature dimensions, effectively stretching or shrinking the manifold along certain axes. Spatial attention re-weights different locations, focusing computational resources on a specific region of the manifold's domain (the image space).

\subsubsection{Lateral Interaction Module (V1)}
\textbf{Implementation:} The V1 module includes a Lateral Interaction block, implemented as a convolution followed by channel-wise self-attention within a residual connection.
\textbf{Geometric Interpretation:} This simulates the horizontal connections within cortical layers that mediate contextual effects like lateral inhibition. Geometrically, this enforces local consistency constraints on the feature manifold. It encourages nearby points in the image space to have related representations, promoting smoothness and helping to form coherent structures like contours, which can be seen as enforcing a local group structure.

\subsubsection{Predictive Coding Loop (AIT to V1)}
\textbf{Implementation:} We implement predictive coding via a top-down connection from the highest level of the ventral stream (AIT) back to V1. The AIT module generates a prediction of V1 feature activations. This prediction is subtracted from the actual bottom-up V1 activity to compute a prediction error, $\epsilon = \text{ReLU}(\text{V1}_{\text{bottom-up}} - \text{AIT}_{\text{top-down}})$. This error signal is then used as an additional learning signal.
\textbf{Geometric Interpretation:} This is the core of our geometric dynamical system. The top-down signal from AIT is a hypothesis about the world, projected back onto the low-level V1 feature manifold. The bottom-up signal is the evidence. The error $\epsilon$ is a vector field on the V1 manifold that drives the learning process, forcing the high-level AIT manifold to generate representations that are consistent with the low-level sensory data. This process refines the entire hierarchy of representational geometries.

\subsubsection{Neuromodulatory Gating}
\textbf{Implementation:} To model the global gain control exerted by neuromodulators, we introduce a Neuromodulation block in key modules (V1, MT, V4). This block applies a learnable, channel-wise multiplicative scaling factor to feature maps.
\textbf{Geometric Interpretation:} This mechanism can be interpreted as controlling the local curvature or metric of the representation manifold. A high gain (scaling factor > 1) could increase the sensitivity of the representation to small input changes (increasing curvature), while a low gain (scaling factor < 1) could promote invariance by flattening the manifold locally. This allows the network to adapt its representational geometry based on context.

\section{Experiments and Results}

We benchmarked VCNet's performance against contemporary neural networks of comparable size to assess its image classification capabilities. We chose datasets and tasks that are particularly relevant to the evolution and function of biological vision and that test the geometric principles embedded in our architecture.

\subsection{Experimental Setup}
All models were trained using the Adam optimizer \cite{kingma2015adam} with a learning rate of $10^{-3}$. We used a batch size of 16 and applied standard data augmentation techniques, including random horizontal flips and random rotations. All experiments were conducted using Google Colab \cite{googlecolab}.

\subsection{Experiment 1: Animal Pattern Classification}

\paragraph{Motivation} Key evolutionary drivers for primate vision include finding food and avoiding predators, tasks that rely heavily on recognizing natural patterns and textures \cite{Kaas2012:Evolution}. The primate visual cortex is thus highly optimized for this purpose. We therefore evaluated our biologically-inspired model on a benchmark focused on classifying animal patterns, which tests the model's ability to learn representations of complex, semi-structured textures.

\paragraph{Methodology} We utilized the Spots-10 dataset, which contains 50,000 grayscale 32x32 pixel images across 10 classes of animal patterns \cite{Atanbori2024:SPOTS10}. We trained VCNet Mini and compared its performance against a suite of established models whose weights were derived via knowledge distillation, making them highly compact and efficient. To ensure a consistent evaluation of convergence, all baseline models were finetuned on Spots-10 for the same number of epochs that VCNet Mini was trained.

\begin{table}[h]
\caption{Test accuracy and model size on the Spots-10 benchmark. Best values are in bold. VCNet Mini demonstrates superior accuracy with significantly fewer parameters, highlighting the efficiency of its architectural priors.}
\label{tab:results}
\centering
\begin{tabular}{lcc}
\toprule
\textbf{Model} & \textbf{Test Accuracy (\%)} & \textbf{Model Size (MB)} \\
\midrule
\textbf{VCNet Mini (Ours)} & \textbf{92.08} & \textbf{0.04} \\
DenseNet121 Distiller & 81.84 & 0.07 \\
ResNet101V2 Distiller & 80.29 & 0.07 \\
ResNet50V2 Distiller & 79.03 & 0.07 \\
MobileNet Distiller & 78.26 & 0.07 \\
MobileNetV3-Small Distiller & 78.04 & 0.07 \\
\bottomrule
\end{tabular}
\end{table}

\paragraph{Results} As shown in Table~\ref{tab:results}, VCNet Mini attains the highest accuracy on Spots-10 (92.08\%), outperforming the strongest baseline (DenseNet121 Distiller, 81.84\%) by a significant margin of 10.24 percentage points. To ensure a fair comparison with the lightweight distilled baselines, we reduced VCNet’s hidden-layer widths to form the \textit{Mini} variant. Remarkably, VCNet Mini achieves this superior performance while using only 0.04~MB of storage, about 43\% smaller than the 0.07~MB baselines. These findings strongly indicate that architectures inspired by the geometric and computational principles of the visual cortex can yield models that are both highly accurate and extremely parameter-efficient.

\subsection{Experiment 2: Light Field Classification}

\paragraph{Motivation} Standard 2D images are flat projections of the 3D world, discarding vast amounts of visual information related to depth, parallax, and view-dependent reflectance. The human visual system (HVS) processes a much richer input, leveraging binocular vision and eye movements to interpret a subset of the 7D plenoptic function \cite{Adelson1991:Plenoptic}. Light field cameras, which capture both the intensity and the angular direction of light rays, provide data that is a much closer analogue to the input processed by the HVS \cite{Lin2024:DeepLearning}. We hypothesize that an architecture designed to emulate the visual cortex's dual-stream, geometrically-aware processing will demonstrate superior performance on this richer data modality.

\paragraph{Methodology} We evaluated VCNet on a light field image classification task using a standard dataset \cite{Raj2016:StanfordLytro}. The light field data was processed into a 4D tensor, which was then fed into the models. We compared its performance against benchmark models: ResNet18, VGG11 with Batch Normalization, and MobileNetV2. These baselines were pre-trained on ImageNet and finetuned for the same number of epochs as VCNet was trained.

\begin{table}[h]
\caption{Performance and Size Comparison on Light Field Image Classification. VCNet achieves the highest accuracy while being the most compact model, demonstrating its suitability for processing higher-dimensional, geometrically rich visual data.}
\label{tab:lightfield_results}
\centering
\begin{tabular}{lcc}
\toprule
\textbf{Model} & \textbf{Test Accuracy (\%)} & \textbf{Model Size (MB)} \\
\midrule
\textbf{VCNet (Ours)} & \textbf{74.42} & \textbf{3.52} \\
MobileNetV2 & 72.09 & 8.66 \\
ResNet18 & 65.12 & 42.69 \\
VGG11\_BN & 51.16 & 491.39 \\
\bottomrule
\end{tabular}
\end{table}

\paragraph{Results} The results, summarized in Table \ref{tab:lightfield_results}, highlight VCNet's superior performance and efficiency. VCNet achieved the highest test accuracy (74.42\%) while maintaining a minimal model size of 3.52 MB. This is over twice as small as MobileNetV2, over ten times smaller than ResNet18, and over 100 times smaller than VGG11. This result validates our hypothesis that an architecture incorporating principles like dual-stream processing and predictive feedback is particularly effective for processing high-dimensional visual data that contains both object identity and spatial/viewpoint information.

\section{Conclusion and Future Work}

In this work, we introduced VCNet, an architecture guided by the computational principles and anatomical organization of the primate visual cortex. By interpreting these principles through the lens of geometry and dynamical systems, framing dual-stream processing as manifold disentanglement and predictive coding as geodesic refinement, we developed a model that demonstrates superior performance and parameter efficiency on challenging image classification tasks. Our findings underscore the significant potential of this approach: by embedding high-level geometric priors into the macro-architecture of a network, we can guide it to learn more robust and efficient representations. This convergence of disciplines not only offers a path toward more capable artificial systems but also provides computational frameworks for testing hypotheses about brain function.

Our model opens several avenues for future research.
\begin{itemize}
    \item \textbf{Integrating Explicit Equivariance:} A powerful next step would be to combine our macro-scale geometric approach with the micro-scale constraints of geometric deep learning. Incorporating steerable filters into the V1 module could enforce explicit rotation and scale equivariance at the lowest level, which could then be integrated into the global representations of the full VCNet architecture.
    \item \textbf{Topological Data Analysis (TDA):} The geometric structure of the learned manifolds could be more formally analyzed using tools from TDA. We could use persistent homology to quantify the topological structure of the representations, testing the hypothesis that better representations have simpler topology (e.g., one connected component per class with no spurious holes).
    \item \textbf{Extension to Spatio-Temporal Dynamics:} The current model is designed for static images. Extending it to video processing is a natural progression. The dorsal stream and recurrent dynamics would become even more critical for modeling the flow of information on spatio-temporal manifolds, potentially leading to more robust action recognition and video prediction models.
    \item \textbf{Geometric Principles in Language:} Finally, it is worth speculating whether the principles of hierarchical, predictive, and geometrically structured representations could be applied to other modalities. Understanding the geometry of representations in large language models is a burgeoning field, and concepts like dual-stream processing (e.g., for syntax vs. semantics) and predictive refinement could offer valuable new architectural ideas.
\end{itemize}

\section*{Author Contributions}

\textbf{Brennen Hill:} Project lead, conceptualization, software, engineering, investigation, research, writing\\
\textbf{Zhang Xinyu:} Software, engineering, investigation, research,writing. \\
\textbf{Timothy Putra Prasetio:} Software, engineering, investigation, research, writing.

\bibliographystyle{plainnat}
\bibliography{main}

\end{document}